\documentclass[
]{ceurart}

\usepackage{amsmath}
\usepackage{algorithm}
\usepackage[noend]{algpseudocode}
\usepackage[T1]{fontenc}
\usepackage{listings}
\usepackage{amssymb}
\usepackage{xcolor}
\usepackage{graphicx}
\usepackage{float}

\usepackage{inconsolata} 

\lstdefinelanguage{json}{
    basicstyle=\ttfamily\footnotesize,
    showstringspaces=false,
    breaklines=true,
    frame=single,
    backgroundcolor=\color{gray!5},
    morestring=[b]",
    literate=
     *{0}{{{\color{blue}0}}}{1}
      {1}{{{\color{blue}1}}}{1}
      {2}{{{\color{blue}2}}}{1}
      {3}{{{\color{blue}3}}}{1}
      {4}{{{\color{blue}4}}}{1}
      {5}{{{\color{blue}5}}}{1}
      {6}{{{\color{blue}6}}}{1}
      {7}{{{\color{blue}7}}}{1}
      {8}{{{\color{blue}8}}}{1}
      {9}{{{\color{blue}9}}}{1}
      {:}{{{\color{black}:}}}{1},
}

\begin{document}

\copyrightyear{2025}
\copyrightclause{Copyright for this paper by its authors.
  Use permitted under Creative Commons License Attribution 4.0
  International (CC BY 4.0).}


\title{Fine-tuning for Better Few Shot Prompting: An Empirical Comparison for Short Answer Grading}
\conference{EvalLAC'25: 2nd Workshop on Automatic Evaluation of Learning and Assessment Content, July 26, 2025, Palermo, Italy}



\author[1]{Joel Walsh}[
  email=jwalsh@ict.usc.edu,
]
\fnmark[1]

\author[2]{Siddarth Mamidanna}[
  email=spmamida@ucsc.edu,
]
\fnmark[1]

\author[1]{Benjamin Nye}[
  orcid=0000-0002-5902-9196
]

\author[1]{Mark Core}[
  orcid=0000-0002-0438-3868
]

\author[1]{Daniel Auerbach}

\address[1]{University of Southern California - Institute for Creative Technologies, Los Angeles, CA USA}
\address[2]{University of California, Santa Cruz, Santa Cruz, CA, USA}

\fntext[1]{These authors contributed equally to this work.}

\maketitle    
\begin{abstract}
Research to improve Automated Short Answer Grading has recently focused on Large Language Models (LLMs) with prompt engineering and no- or few-shot prompting to achieve best results. This is in contrast to the fine-tuning approach, which has historically required large-scale compute clusters inaccessible to most users. New closed-model approaches such as OpenAI’s fine-tuning service promise results with as few as 100 examples, while methods using open weights such as quantized low-rank adaptive (QLORA) can be used to fine-tune models on consumer GPUs. We evaluate both of these fine-tuning methods, measuring their interaction with few-shot prompting for automated short answer grading (ASAG) with structured (JSON) outputs. Our results show that finetuning with small amounts of data has limited utility for Llama open-weight models, but that fine-tuning methods can outperform few-shot baseline instruction-tuned LLMs for OpenAI's closed models. While our evaluation set is limited, we find some evidence that the observed benefits of finetuning may be impacted by the domain subject matter. Lastly, we observed dramatic improvement with the LLama3.1 8B-Instruct open-weight model by seeding the initial training examples with a significant amount of cheaply generated synthetic training data. 
\end{abstract}


%

\section{Introduction}
The widespread adoption of Massive Open Online Courses (MOOCs) and Learning Management Systems has created an increasing amount of learning assessments in online, machine-readable formats. Due to the volume of assessment responses and sometimes teacher-less nature of these courses, Automated Short Answer Grading (ASAG) has become a robust research target. Most attempts have used few or no shot learning with baseline LLMs.  The practice of fine-tuning LLMs offers some promise as a way to increase the performance of LLMs on specific tasks, but the cost of compute and data collection budget are significant constraints. This paper explores finetuning under realistic constraints. Specifically, given a modest set of labeled examples (N=148) and a single‑GPU training envelope, when does parameter‑efficient fine‑tuning (QLoRA or OpenAI’s in‑house tuning) yield statistically and practically meaningful gains over few‑shot prompting for multi‑concept ASAG? We find that the answer to this question is somewhat nuanced, as some methods of fine-tuning modestly improve ASAG in this context, while other methods do not.

Rather than train the models on a specific content domain, this fine-tuning approach trains across a varied set of domains and with different numbers of few-shot prompts, with the goal to tune a model that uses few-shot prompts more effectively for new content areas. With this specific task the model must assign binary labels for demonstrated understanding of user-defined concepts, or learning objectives. The evaluation set consists of expert human-graded responses from different domain areas. This particular type of structured JSON output is particularly useful as agents and multi-agent systems have shown their usefulness at linking LLM outputs to software decisions.

\subsection{Related Work} 
Deep learning for automated text scoring has been the focus of numerous public dataset challenges, but the approaches that have come from these competitions often focus on essay-length text and often ignore issues such as textual entailment \cite{Alik}. Since the widespread adoption of neural methods in Natural Language Processing, there have been many attempts to utilize the latest network architectures for Short Answer Grading (SAG). These approaches included using LSTMs \cite{kumar2017}, mixing sentence level and token level embeddings  \cite{saha2018}, and a variety of transformers-based approaches \cite{liu2019} \cite{wang2019}. 

As LLMs began to perform well on benchmark tasks similar to ASAG, such as Question Answering \cite{brown2020}, several groups began to experiment with leveraging the no-shot and few-shot capabilities of LLMs for ASAG \cite{yoon} \cite{schneider2024} \cite{ivanova}. Since the release of industry open weight models (e.g., Meta's Llama family), there has been limited research comparing ASAG performance with fine-tuned closed models (i.e., different sizes of Open AI's GPT-4) \cite{chamieh2024}. 

Supervised finetuning (SFT) and Instruction-tuned models \cite{wei2019} like ChatGPT have no doubt changed the world in the past few years; but training often requires a lot of gas (i.e., large numbers of examples). In the case of SFT, optimal prompt-response pairs is the gas. Since the proliferation of instruction tuned models, researchers have had success mitigating this need by using existing human-annotated data to drive reinforcement learning \cite{chen2024} and using high-quality examples as a seed to create additional SFT data \cite{zhu2025}.

Although our approach to these empirical studies on ASAG are similar, we expand these comparisons in a few areas. In this study, we also fine-tune a large open weight model for ASAG. We feel that this is an important departure as closed models do not disclose details about the parameter count, hyperparameters, or techniques used in fine-tuning. As many organizations are compute-limited, we focused on training quantized 4-bit models that can fit on one NVIDIA A40 GPU. We also study the effect of varying the amount N of few shot examples (i.e., N-shot) used to prompt each model at test time. We focus on producing JSON-structured outputs, which play an important role in agentic and LLM-based software. Lastly, we seed synthetic data using a small number of examples created by real subject matter experts and learners. This approach will enable low-resource educational organizations to create supervised finetuning data on a scale that makes the technology viable.

\section{Methods}

\subsection{Fine-tuning and baseline models}
For the closed models, we used OpenAI's GPT-4o-mini. OpenAI does not publish parameter counts for either of these models, or key information about the architecture. OpenAI has offered fine-tuning services starting in August of 2023\cite{fine_tuning_api}. The documentation for the fine-tuning API states that, "We typically see improvements from fine-tuning on 50 to 100 training examples" \cite{openai}. However, OpenAI does not allow you to download the fine-tuned models, or any of their models.

For the open models, we used LLama3.1 8B-Instruct \cite{gratt2024}. Like GPT-4o-mini, this model is already instruction-tuned. The model size is 8 billion parameters, which is the largest model that can be finetuned on one NVIDIA A40 GPU with 48 GB of RAM using a Quantized Low Rank Adapter (QLoRA) \cite{dettmers2023} approach. Each model was evaluated on a set of 148 distinct short-answer labeled examples, which was effectively extended to 17820 prompts (Sec.~\ref{sec:eval-data-generation}), which took approximately 40-50 hours to run on our university's NVIDIA A40 GPUs. Due to these compute limitations at test time we were not able to run extensive ablation tests on model architecture, or on hyperparameters like LoRa rank or number of hidden dimensions. 

\subsection{Data collection}
Data was gathered with permission from the OpenTutor project \cite{nye}. OpenTutor allows curriculum creators to author tutoring dialogues, where assessment of concept understanding is conducted via multi-concept ASAG on student responses. Subject matter experts created the dialogues, and also graded student responses in order to improve the system. The subject matter of the training set pulled from technical subjects for which the models certainly had exposure (biology and computer science) to subjects where the models likely had much less exposure; or where best-practices may be changed over time, such as military leadership or culture. Each authored dialogue contains several concepts, or key points that the responses were supposed to address. For example, a dialogue on invasive species contains three concepts:
\begin{enumerate}
    \item Monitoring and collecting data on the invasive species is key to fighting them,
    \item Rapid or a quick response is another key to fighting invasive species, and 
    \item AI and machine learning can be used to find and track the spread of invasive species faster.
\end{enumerate}

When a subject matter expert grades the answer, they provide an either 0 or 1 label for each concept; a 0 if the answer does not demonstrate knowledge of the concept and a 1 if it does. Some answers contain correct statements about one concept without mentioning the others; this is reflected in the grades as a skipped concept. 

All Short Answer questions and training examples were generated as a byproduct of user interaction with the OpenTutor dialogue-based tutoring system. The corpus of graded responses for this analysis comes from multiple studies conducted with adult learners, to include both students and Mechanical Turk workers. The MTurk workers were screened for appropriate study behavior (e.g., enough time spent, expert human review of answers) Subject matter experts were then asked to grade each answer, specifying whether or not it addressed the concepts dictated by the expert.

\subsection{Fine-Tuning Training Data}

Subject matter experts labeled only a portion of user responses from OpenTutor dialogues; only labeled responses were used in the current research. The training data included all \textit{labeled} responses from 42 lessons, excluding three lessons that were set aside for evaluation. Lessons without graded responses were omitted.

Due to varying levels of use and grading by experts, each lesson had a different number of graded responses (ranging from as few as 4 to over 100). To generate training data, multiple subsets of graded responses were randomly selected from each lesson, with each subset containing between 1 and 40 examples. Each subset was then individually paired with another distinct graded response from the same lesson, forming multiple complete training examples. Each of these pairings represented an \textit{n}-shot example, where \textit{n} denotes the number of responses in the subset.

Initially, the graded responses contained only binary labels (true/false) for each concept, without confidence scores or justifications. To enhance the quality and informativeness of the data, we used GPT-4o to generate justifications and assign confidence scores for each concept. Specifically, each concept label within a response was individually passed to GPT-4o along with the corresponding student response, generating a justification and confidence rating. To better calibrate the confidence ratings provided by GPT-4o, we averaged the confidence scores with those obtained from an existing logistic regression classifier previously trained on the same dataset \cite{nye2021}. This resulted in a more accurate reflection of the true confidence. All samples (a total of 148 training examples) were manually reviewed afterward to ensure their quality and correctness.

\vspace{1em}

\subsection{Generating Synthetic data}

To extend this research beyond this small amount of hand-labeled examples, we also investigated the impact of training LLama3.1 8B-Instruct using synthetic data. To generate synthetic data, the training data was split into a 90:10 train-validation split. Google Gemini's 1.5 Flash model was used for generation. Our process involved randomly choosing 1-3 examples from either the train or validation, and appending them to a prompt to ``generate one additional example from an academic, corporate, or military training domain''. If the example did not form a valid JSON, it was thrown out. These one thousand examples fortified the existing training and validation sets. The test set remained entirely real data. For 1.5 Flash, this process cost 45 cents (US), offering a relatively cost-effective process. Surprisingly the latest Gemini thinking model, 2.5 Flash, could not create valid JSONs with any regularity. A limitation of this study is that the amount of synthetic data was chosen somewhat arbitrarily, as we did not have the test-time compute resources to test models trained on several different amounts of synthetic data. Also, we were unable to fine-tune GPT 4o-mini with synthetic data to see how it would affect GPT 4o-mini. This will be explored in future work.
 
\subsection{Generating Evaluation Data}\label{sec:eval-data-generation}

Evaluation data consisted of three “gold” lessons withheld from training, each covering a distinct topic:

\begin{enumerate}
    \item \textbf{Diode Breakdown} (technical)
    \item \textbf{Reaching Out} (leadership-focused military context)
    \item \textbf{Suicide Prevention} (general risk factor awareness)
\end{enumerate}

Each gold lesson had manually graded responses to ensure high evaluation quality: Diode Breakdown (50 responses), Reaching Out (50 responses), and Suicide Prevention (120 responses).

\begin{figure}
    \centering \includegraphics[width=0.75\linewidth]{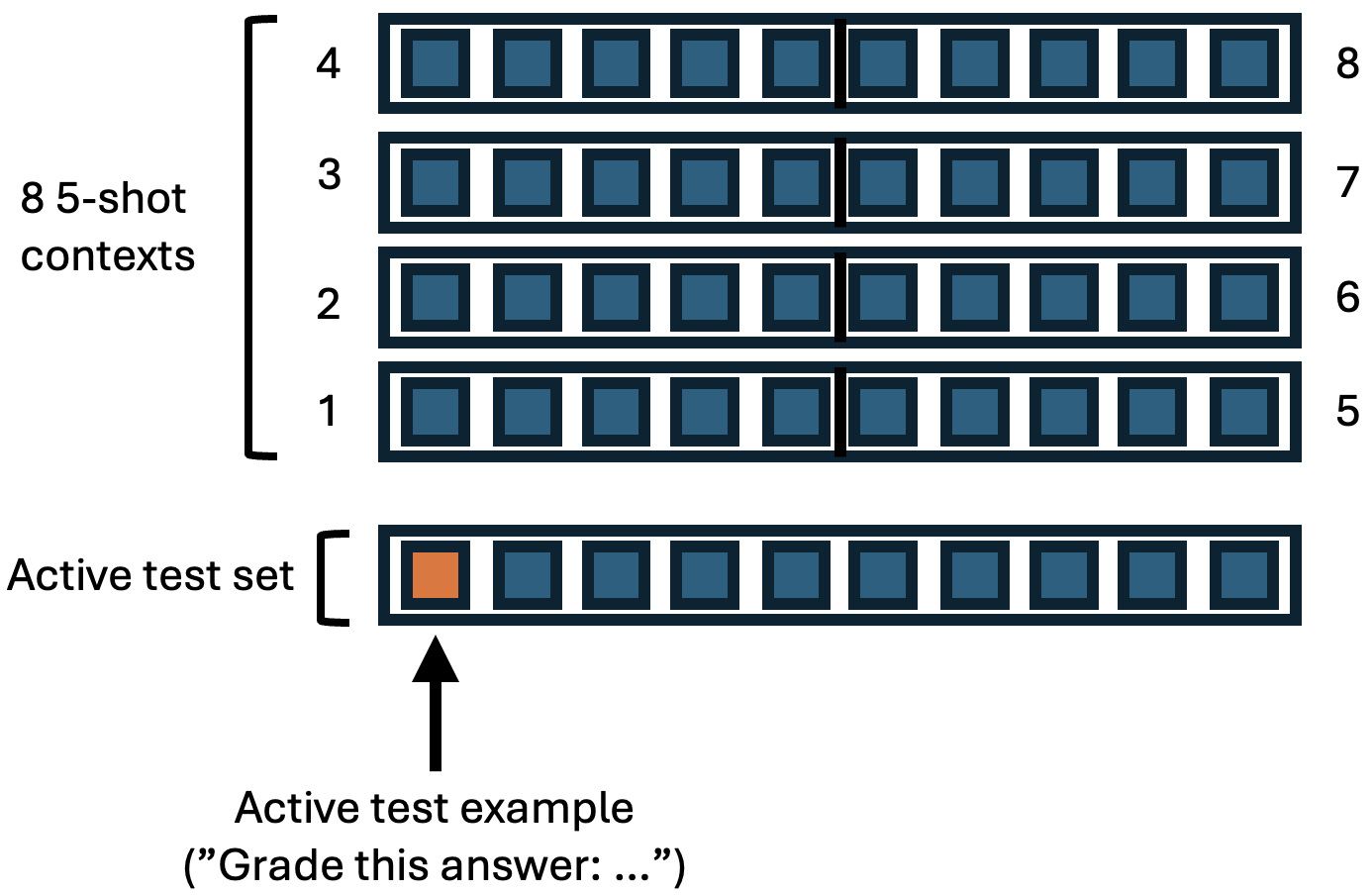}
    \caption{Visualization of evaluation structure for N=5 and 50 examples. All examples in the active test set consist of only the answer, while the n-shot examples contain the graded answers.}
    \label{fig:eval-diagram}
\end{figure}

To simulate realistic scenarios of incremental data annotation and model updating, we evaluated the system using an N-shot strategy at different numbers of examples. For example, when N=5, five graded responses were appended to the prompt as context examples. To ensure robust evaluation despite limited available data, we structured the evaluation as follows (see Fig.~\ref{fig:eval-diagram}):

\begin{enumerate}
\item \textbf{Chunking.} Responses from each gold lesson were grouped into test sets containing 10 responses each. This followed an approach similar to cross-validation, so for a lesson with 50 responses, there were 5 test sets.
\item \textbf{Creating N-shot Examples.} For each test set, multiple N-shot contexts were created by drawing graded examples from the remaining responses within that lesson. For instance, if a lesson contained 50 responses grouped into 5 test sets (each containing 10 items), each test set would have 8 distinct N-shot context sets. The total number of N-shot sets was calculated as (50 - 10) / 5 = 8 sets. This process was repeated for N=0, N=5, N=10, ..., and while the N-shot contexts sometimes overlapped with each other, test items never overlapped with any N-shot examples. Consequently, each test response was evaluated multiple times, each time across different values of N and N-shot contexts.
\item \textbf{Multiple Trials for Robustness.} When N > 0, we generated multiple evaluation trials (typically 10 trials per set) to ensure stability and reliability in our assessment. For each trial, we reshuffled the selected N-shot examples, not just altering their order but also simulating realistic variability by randomly ablating certain labeled concepts. This meant that, across trials, some contexts were richly annotated, while others were sparsely annotated—mimicking real-world annotation inconsistencies.
\item \textbf{Comprehensive Evaluation.} Every response in the test sets was evaluated across all trials and corresponding N-shot contexts. This extensive evaluation approach ensured a thorough and robust assessment, enabling confident observations and reliable conclusions. As we mentioned earlier the n-shot prompts also made evaluation somewhat slow (approximate 50 hours) for the open weight models, as self-attention inference scales quadratically with context window length.
\end{enumerate}

Ultimately, this expanded our final test size across all lessons to be \textbf{n=17820 examples}, providing a rigorous and thorough evaluation procedure.

\section{Results}


Our analysis finds that, for this particular type of structured output, fine-tuning GPT 4o-mini on just $\sim 150$ examples was impactful, resulting in an F1 increase from 0.68 to 0.73. The largest improvement occurs in the Reaching Out and Suicide Prevention domains, which are based on highly specific content domains and whose evaluations require using labeled examples rather than prior knowledge. The objective of fine-tuning the model on a series of few-shot examples is to allow it to adapt far quicker to the grading criteria and knowledge in the few-shot examples, as opposed to relying on prior knowledge and idiosyncratic judgment. This alignment is not just about boosting raw scores; it also makes the model’s output more interpretable and consistent with human feedback loops. By internalizing each specific lesson's grading framework, the fine-tuned model better generalizes to new topics with lower shot counts, delivering reliable, rubric-compliant assessments.

\begin{table}[htbp]
\centering
\caption{Metrics for Different Baseline and Fine-Tuned Model Configurations}
\begin{tabular}{l@{\hspace{1em}}c@{\hspace{1em}}c@{\hspace{1em}}c@{\hspace{1em}}c}
\hline
\textbf{Model} & \textbf{Accuracy} & \textbf{Precision} & \textbf{Recall} & \textbf{F1 Score} \\
\hline
Base GPT-4o-mini & \textbf{0.812} & \textbf{0.936} & 0.550 & 0.681 \\
Fine-Tuned GPT-4o-mini & 0.789 & 0.749 & \textbf{0.741} & \textbf{0.735} \\
Baseline Llama-3-8B & 0.587 & 0.493 & 0.106 & 0.165 \\
Fine-tuned Llama-3-8B (3 Epochs) & 0.586 & 0.287 & 0.046 & 0.078 \\
Fine-tuned Llama-3-8B (6 Epochs) & 0.571 & 0.459 & 0.373 & 0.408 \\
+ (w/ Gemini 1.5 synthetic training data) & 0.721 & 0.663 & 0.650 & 0.653 \\
Fine-tuned Llama-3-8B (9 Epochs) & 0.554 & 0.435 & 0.336 & 0.376 \\
\hline
\end{tabular}
\label{tab:model-metrics}
\end{table}

\begin{figure}
\includegraphics[width=\textwidth]{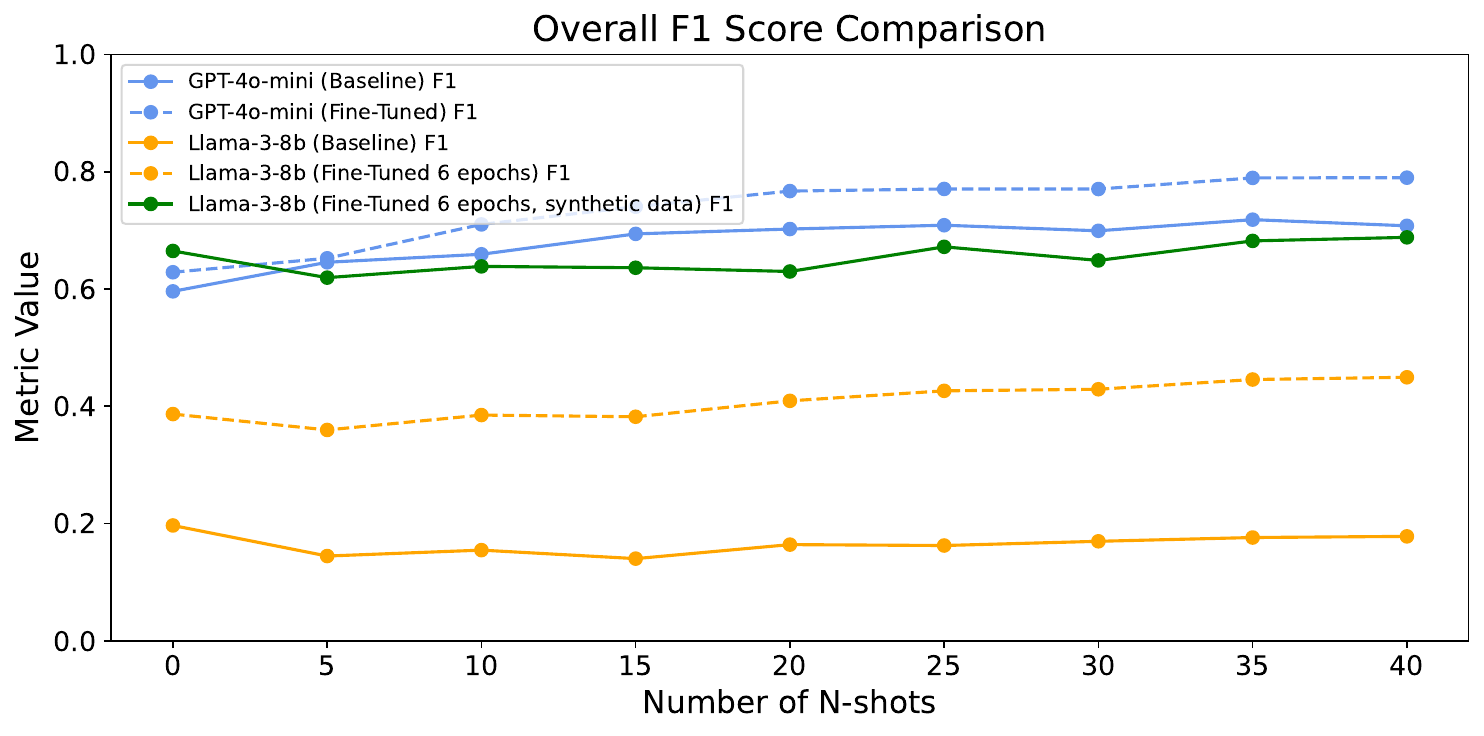}
\caption{F1 scores for baseline and fine-tuned models (Using best Llama model, 6 epoch)} \label{ovr-f1-comp}
\end{figure}

\begin{figure}
\centering
\includegraphics[width=\linewidth]{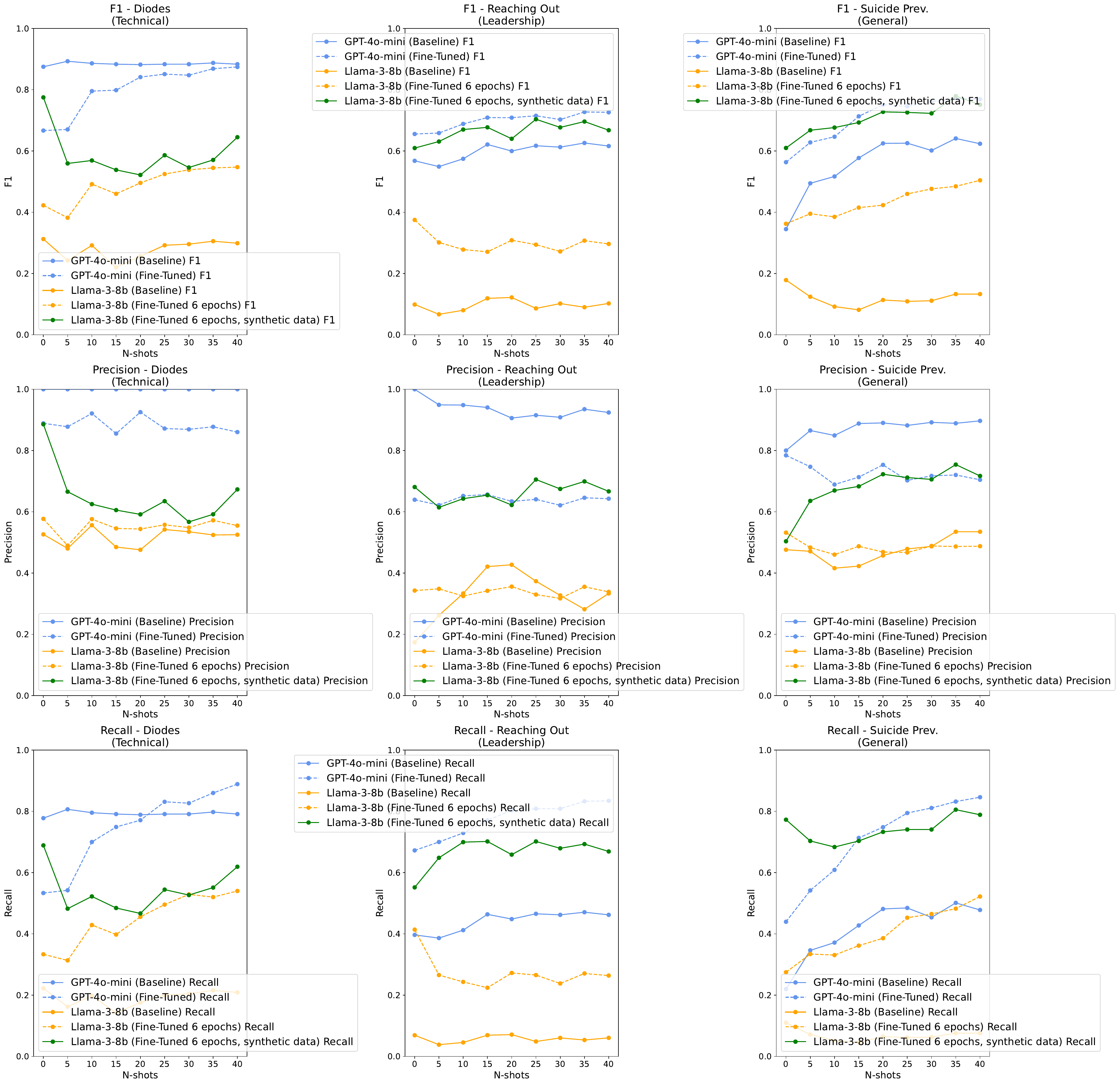}
\caption{Performance metrics across different text domains within the training data } \label{combined-graph}
\end{figure}

Additionally, the fine-tuned GPT 4o-mini model's F1 score improved as the number of N-shots increased. The F1 score graph for the Diodes lesson is impressive (Fig.~\ref{combined-graph}), as it shows that while the base model had a higher initial F1, it did not improve with increasing N-shots. However, the F1 for the fine-tuned model started significantly lower than the base and progressively improved with the number of N-shots, eventually matching the performance of the base model. Looking at the overall F1 score comparison (Fig.~\ref{ovr-f1-comp}) also displays the fine-tuned model adapting to new data more effectively than the base model, as the F1 score increases at a slightly higher rate as the number of N-shots increase for the fine-tuned model.

On the other hand, fine-tuning LLama3.1 8B-Instruct using QLORA on the initial data was not as successful. The baseline and 1 epoch fine-tuned model sometimes failed to generate stopping points, entered phases of repeating, and mainly predicted one class - False. While the model did show some signs of life at epoch 6, the F1 score was still only 0.408. This still however, represents a large gain on the baseline model (see Table~\ref{tab:model-metrics}). The fine-tuning (for 6 epochs) ultimately provided a significant increase in performance over the base model. At 9 epochs, the model's performance began to degrade, suggesting that the threshold for overfitting had been reached. Adding in the synthetic data caused the most dramatic improvement. The best model in both instances was 6 epochs, and boosted the F1 score from 0.408 to 0.653, nearly matching the baseline GPT 4o-mini model. This would suggest that synthetic data can be incredibly effective, and also that this family of models requires much more data than 50-100 examples in order to grasp structured ASAG tasks like this.

\section{Discussion and Future Research}

While the Llama QLORA fine-tuning did not work as well as the GPT 4o mini fine-tuning, there are some major advantages to tuning and serving open-weight models. For one, OpenAI will never let a user download the model weights or architecture. Any system built on OpenAI's technology will have vendor lock-in, and a dependence on an internet connection. LLama3.1 8B-Instruct with synthetic data was close to GPT 4o-mini performance and trending enough in a positive direction to justify serving this model as a viable alternative.

Recent research into structured LLM outputs has shown that constraining LLMs to structured outputs can have a deleterious effect on model reasoning and domain knowledge capabilities \cite{tam2024}. Future research will explore using an agentic process, whereby a model is prompted to provide a free-form justification and a grade, and then a agent is prompted to parse the information into a JSON using some combination of handwritten rules or additional LLM calls. Building learning engineering systems with structured outputs is a constant battle with the noisy output of LLMs. Agents built on handwritten rules or additional LLM calls can also serve as means to tweak "almost there" responses that might be off by a data type, or a slightly incorrect key value.

Other levers for improvement include prompt complexity and the size of 
both synthetic and real training datasets. The real dataset required content creation, human short answers, and a round of subject matter expert grading. This is time consuming and expensive to collect. As an alternative to collecting thousands of real training examples at great cost or generating synthetic data, a potential alternative could be to use the Deepseek approach of ``cold start'' supervised fine-tuning followed by Group Relative Policy Optimization (GRPO) reinforcement learning to further refine the model \cite{deepseek2025}. This process is thought to be much more sample efficient than supervised finetuning.
 
In terms of training with synthetic data, this process can optimized substantially. The prompt could be optimized, the amount of data could be increased, and much more advanced models than Gemini 1.5 Flash now exist. However, the iterative nature of data annotation could lend itself quite well to simulation by multi-agent system, where teacher and student agents with different profiles, prior knowledge, and directives work in conjunction to create new examples. This could create higher-quality synthetic examples.

\section{Conclusion}

This study explored realistic approaches to supervised fine-tuning of LLMs for automated short answer grading tasks. Our findings have shown that fine-tuning large commercial models (such as GPT-4o-mini) with relatively small amounts of data enhances performance over baseline few-shot prompting methods, even in highly context-dependent domains. Meanwhile, QLORA finetuning LLama-3-8B models initially showed poor performance for this particular task, but additional infusions of synthetic data made them competitive with GPT models. 

Overall, the effectiveness of this synthetic data-infused training approach holds great promise for developing reliable ASAG systems that can function without relying on large corporations, server-scale GPUs, or the presence of an Internet connection. Continued exploration into synthetic data generation and hybrid agentic pipelines suggest further performance gains. Collectively, these approaches can enable resource-limited educational organizations to deploy and use strong ASAG technologies, helping democratize access to reliable AI-driven assessment methods.



\appendix
\section*{Appendix}

\section{Prompt template, N=30}
\label{sec:appendix_exp}
\begin{lstlisting}[language=json, caption={Abridged sample of 30-shot prompt}]
{

The user provided an answer to a tutoring question. The answer is provided in JSON format. You are a tutor who is evaluating if the answer is sufficient to show that the user knows a one or more "concepts" which will be labeled "concept_1" to "concept_N". For each concept you evaluate, you must also express how confident you are in your evaluation of how well the answer shows knowledge of each concept. You will also provide a brief justification.

These are the concepts for this lesson.
{
  "concept_1": "They're much less likely to commit suicide now or in the future, because suicidal urges last less than an hour.",
  "concept_2": "The person is still at risk compared to other people, because they still have suicidal thoughts."
}
This is my answer provided in JSON to be evaluated.
{
  "answer_text": "it lowers"
}
Please respond in the following format:
{
  "answer": {
    "answer_text": "string // State the text of the particular answer being classified.",
    "concepts": {
      "concept_N": {
        "is_known": "string // true or false. If the input answer implies that the concept is known, the classification should be true. Otherwise it should be false.",
        "confidence": "float // A 0 to 1 score indicating certainty that a classification is correct. Confidence scores closer to 1 represent higher certainty, and confidence scores closer to 0 represent lower certainty.",
        "justification": "string // Why you believe the user answer is or is not sufficient to determine if they know the concepts."
      }
    }
  }
}
Only respond with the JSON output in the exact format of the template and no other words or symbols. The output must be valid JSON. Check that the output is valid JSON. 

Here are some examples that have already been labeled (although they may not be fully labeled). They are presented in JSON format, where the answer is given, followed by a concept and a true or false label. Consider these to be ground truth examples.
{
  "answer_1": {
    "answer": "It'll deter them from wanting to commit suicide.",
    "concept_1": "true"
  },
  "answer_2": {
    "answer": "they're still more likely as they have suicidal ideation",
    "concept_1": "false"
  },
  .
  .
  .
  "answer_30": {
    "answer": "they may not have anything else to use",
    "concept_1": "false"
  }
}
\end{lstlisting}


\section{Synthetic Content Generation Algorithm}

\begin{algorithm}[H]
\caption{Synthetic Content Generation Algorithm}
\label{alg:content_generation}
\begin{algorithmic}[1]

\Function{GeneratePrompt}{\textit{few\_shot\_examples}}
    \State \textit{prompt} $\gets$ \parbox[t]{0.7\linewidth}{"You are an expert AI assistant..."}
    \State \textit{prompt} $\gets$ \textit{prompt} + \parbox[t]{0.7\linewidth}{"Your goal is to create new examples..."}
    \State \textit{prompt} $\gets$ \textit{prompt} + \parbox[t]{0.7\linewidth}{"Please generate ONE new example..."}
    \State \textit{prompt} $\gets$ \textit{prompt} + \parbox[t]{0.7\linewidth}{"The content of the new example should belong to..."}
    \State \textit{prompt} $\gets$ \textit{prompt} + \parbox[t]{0.7\linewidth}{"Ensure the generated examples are distinct..."}
    \State \textit{prompt} $\gets$ \textit{prompt} + \parbox[t]{0.7\linewidth}{"Here are some examples to learn from..."}
    \State \textbf{for each} \textit{example} \textbf{in} \textit{few\_shot\_examples} \textbf{do}
        \State \textit{prompt} $\gets$ \textit{prompt} + \parbox[t]{0.6\linewidth}{"\textbackslash n--- Example ---\textbackslash n" + \textsc{ToJson}(\textit{example})}
    \State \textit{prompt} $\gets$ \textit{prompt} + \parbox[t]{0.7\linewidth}{"\textbackslash nNow, generate a new, unique example..."}
    \State \textbf{return} \textit{prompt}
\EndFunction

\Function{GenerateFromModel}{\textit{model}, \textit{prompt}}
    \State \textit{response} $\gets$ \parbox[t]{0.7\linewidth}{\textit{model}.\textsc{generate}(\textit{prompt}, \textsc{GenerationConfig}, \textsc{SafetySettings})}
    \State \textit{json\_object} $\gets$ \textsc{ExtractJson}(\textit{response}.\textsc{text})
    \State \textbf{return} \textit{json\_object}
\EndFunction

\Procedure{Main}{}
    \State \textit{originalExamples} $\gets$ \textsc{LoadExamples}(\textsc{InputFile})
    \State \textsc{WriteExamples}(\textsc{OutputFile}, \textit{originalExamples})
    \State \textit{generatedCount} $\gets$ 0
    \State \textit{failedAttempts} $\gets$ 0
    \State \textit{model} $\gets$ \textbf{new} \textsc{GenerativeModel}(\textsc{ModelName})
    \While{\textit{generatedCount} < \textsc{TargetCount} \textbf{and} \textit{failedAttempts} < \textsc{MaxFailedAttempts}}
        \State \textit{k} $\gets$ \textsc{RandomInt}(1, 3)
        \State \textit{fewShotSamples} $\gets$ \textsc{RandomSample}(\textit{originalExamples}, \textit{k})
        \State \textit{promptText} $\gets$ \textsc{GeneratePrompt}(\textit{fewShotSamples})
        \State \textit{newExample} $\gets$ \textsc{GenerateFromModel}(\textit{model}, \textit{promptText})
        \If{\textit{newExample} is valid}
            \State \textsc{AppendExample}(\textsc{OutputFile}, \textit{newExample})
            \State \textit{generatedCount} $\gets$ \textit{generatedCount} + 1
            \State \textit{failedAttempts} $\gets$ 0
        \Else
            \State \textit{failedAttempts} $\gets$ \textit{failedAttempts} + 1
        \EndIf
    \EndWhile
    \State \textsc{Print}(\parbox[t]{0.7\linewidth}{"Process finished. Total generated: " + \textit{generatedCount}})
\EndProcedure

\end{algorithmic}
\end{algorithm}

\begin{acknowledgments}
    
 The project or effort depicted was or is sponsored by the U.S. Government under contract number W912CG-24-D-0001 and W911NF-14-D-0005, as part of the USC ICT UARC and the AI Research Center of Excellence in Education (AIRCOEE). The content of  the  information  does  not  necessarily  reflect  the  position  or  the  policy  of  the  Government,  and  no  official endorsement should be inferred.
\end{acknowledgments}
\section*{Declaration on Generative AI}

  The author(s) have not employed any Generative AI tools.
  \newline


%
%
%
%

\clearpage
\bibliography{sample-ceur}

\end{document}